\documentclass{article}

    \usepackage[preprint]{neurips_2025}

\usepackage[utf8]{inputenc} 
\usepackage[T1]{fontenc}    
\usepackage{hyperref}       
\usepackage{url}            
\usepackage{booktabs}       
\usepackage{amsfonts}       
\usepackage{nicefrac}       
\usepackage{microtype}      
\usepackage{xcolor}         

\usepackage{algorithm,algorithmic}  
\usepackage{amsmath,amssymb,amsthm}
\usepackage{multirow}
\usepackage{graphicx}
\usepackage{wrapfig}
\usepackage{subfigure}
\usepackage{pifont}

\usepackage[most]{tcolorbox}
\newtcolorbox[auto counter]{takeaway}[1][]{
    colback=gray!20, 
    colframe=gray!80,
    boxrule=0.5mm, 
    enhanced, 
    before upper={\textbf{Takeaway~\thetcbcounter:}~},
    #1
}

\title{Exploring Federated Pruning for Large Language Models}

\author{Pengxin Guo$^{1}$ Yinong Wang$^{1}$ Wei Li$^{2}$ Mengting Liu$^{3}$ Ming Li$^{4}$ Jinkai Zheng$^{5}$ Liangqiong Qu$^{1}$\thanks{Corresponding author.}~~
\\
$^{1}$The University of Hong Kong \quad $^{2}$Southern University of Science and Technology \\
$^{3}$Sun Yat-sen University \quad $^{4}$Guangming Laboratory \quad $^{5}$Hangzhou Dianzi University \\
\texttt{\{guopx,u3011648\}@connect.hku.hk, li.wei.ml.619@gmail.com,}\\
\texttt{liumt55@mail.sysu.edu.cn, ming.li@u.nus.edu,} \\
\texttt{zhengjinkai3@hdu.edu.cn, liangqqu@hku.hk} \\
}

\begin{document}

\maketitle

\begin{abstract}
  LLM pruning has emerged as a promising technology for compressing LLMs, enabling their deployment on resource-limited devices. However, current methodologies typically require access to public calibration samples, which can be challenging to obtain in privacy-sensitive domains. To address this issue, we introduce \textbf{FedPrLLM}, a comprehensive federated pruning framework designed for the privacy-preserving compression of LLMs. In FedPrLLM, each client only needs to calculate a pruning mask matrix based on its local calibration data and share it with the server to prune the global model. This approach allows for collaborative pruning of the global model with the knowledge of each client while maintaining local data privacy. Additionally, we conduct extensive experiments to explore various possibilities within the FedPrLLM framework, including different comparison groups, pruning strategies, and the decision to scale weights. Our extensive evaluation reveals that one-shot pruning with layer comparison and no weight scaling is the optimal choice within the FedPrLLM framework. We hope our work will help guide future efforts in pruning LLMs in privacy-sensitive fields. Our code is available at \href{https://github.com/Pengxin-Guo/FedPrLLM}{https://github.com/Pengxin-Guo/FedPrLLM}.

\end{abstract}

\section{Introduction}
\label{sec:intro}

Large Language Models (LLMs) \cite{brown2020language,touvron2023llama,achiam2023gpt} have revolutionized the field of natural language processing by demonstrating remarkable capabilities across various tasks. However, their increasing size leads to significant hardware requirements, limiting real-world deployment. To address this, research has focused on compact LLMs through compression techniques, such as \textit{pruning} \cite{ma2023llm,frantar2023sparsegpt,sun2024a}, \textit{knowledge distillation} \cite{gu2024minillm,xu2024survey}, \textit{quantization} \cite{xiao2023smoothquant,shao2023omniquant}, and \textit{low-rank factorization} \cite{zhao2024galore,saha2023matrix}. Among these, pruning has emerged as a promising method to reduce resource demands by selectively removing redundant parameters while preserving performance \cite{ma2023llm,frantar2023sparsegpt}. Typically, LLM pruning methods can be broadly classified into \textit{structured pruning}, which removes entire substructures within LLMs, such as neurons \cite{ma2023llm,li2023losparse,ashkboos2024slicegpt}, layers \cite{xia2024sheared}, or even entire transformer blocks \cite{gromov2025the}, and \textit{unstructured pruning}, which removes individual weights from the model's weight matrices based on certain criteria \cite{frantar2023sparsegpt,sun2024a,zhang2024plug,yin2024outlier,xu2024besa}. This work focuses on unstructured pruning, as it tends to achieve higher compression rates and maintain better model performance compared to structured pruning \cite{frantar2023sparsegpt,he2024matters,xia2024sheared,zhang2024plug}.

Despite advances in LLM unstructured pruning methods, these approaches usually rely on access to public calibration data to guide the pruning process \cite{frantar2023sparsegpt,sun2024a,zhang2024plug,yin2024outlier,xu2024besa}. Specifically, they require calibration samples to evaluate the importance of the model weights in order to determine the pruning mask matrix for pruning models. 
However, in many real-world scenarios, such as healthcare, finance, and personalized services, the data used for pruning might be private and cannot be shared due to privacy regulations and concerns.
Federated Learning (FL) \citep{mcmahan2017communication,zhang2024flhetbench,zeng2024tackling,guo2025new,guo2025exploring}, which utilizes collaborative and decentralized training of models across multiple institutions without sharing personal data externally, offers a promising solution to this challenge.

Integrating FL with LLM pruning allows each client to calculate a local pruning mask matrix based on its private calibration data and share it with the server. The server then aggregates these mask matrices into an aggregated mask matrix and selects the top-k values (the most clients want to prune) to derive a final pruning mask matrix for pruning the global model. Despite its ability to protect data privacy, three unresolved challenges within this framework hinder practical deployment.

\textbf{Challenge 1: How to compare parameters? } 
When selecting the top-k values, a critical ambiguity arises: Should parameter importance be compared across the entire layer or within each respective row or column (corresponding to \textit{layer}, \textit{row}, and \textit{column comparisons}, respectively)? 
Previous centralized LLM pruning work \cite{sun2024a} has highlighted the importance of using a proper comparison group for pruning LLMs, yet no study explores this in federated scenarios.

\textbf{Challenge 2: To scale or not scale for retained parameters.} 
Beyond simply determining which parameters to prune via majority voting (i.e., selecting top-k values), the FL aggregated mask matrix reveals a critical hidden signal: how strongly each parameter is disfavored across clients. Consider two surviving parameters - one narrowly retained (pruned by 10/100 clients) and another unanimously preserved (pruned by 0/100 clients). Traditional pruning treats both equally, maintaining their original magnitudes despite their differing consensus levels. However, this ignores a critical insight: the former parameter, though retained, exhibits weaker consensus across clients. This observation raises a fundamental question: Rather than simply employing binary masking, could we leverage the FL aggregated mask matrix to guide continuous weight adjustment, where retained parameters are scaled down proportionally based on their pruning frequency?


\textbf{Challenge 3: Is iterative pruning worth the cost?} LLM pruning is typically performed \textit{layer-by-layer} recursively to avoid error accumulation \cite{frantar2023sparsegpt,sun2024a,zhang2024plug}. As a result, in FL, this necessitates either \textit{one-shot pruning} (clients compute all layer mask matrices and share them with the server in one go) or \textit{iterative pruning} (clients send the mask matrices to the server layer by layer in an iterative manner).
While iterative pruning allows for refining the local model promptly, it incurs prohibitive communication costs for deep LLMs.
This raises an unstudied question: Does iteratively refining the local model improve accuracy enough to justify its massive communication overhead?

\begin{figure} [t]
    \centering
    \includegraphics[width=\linewidth]{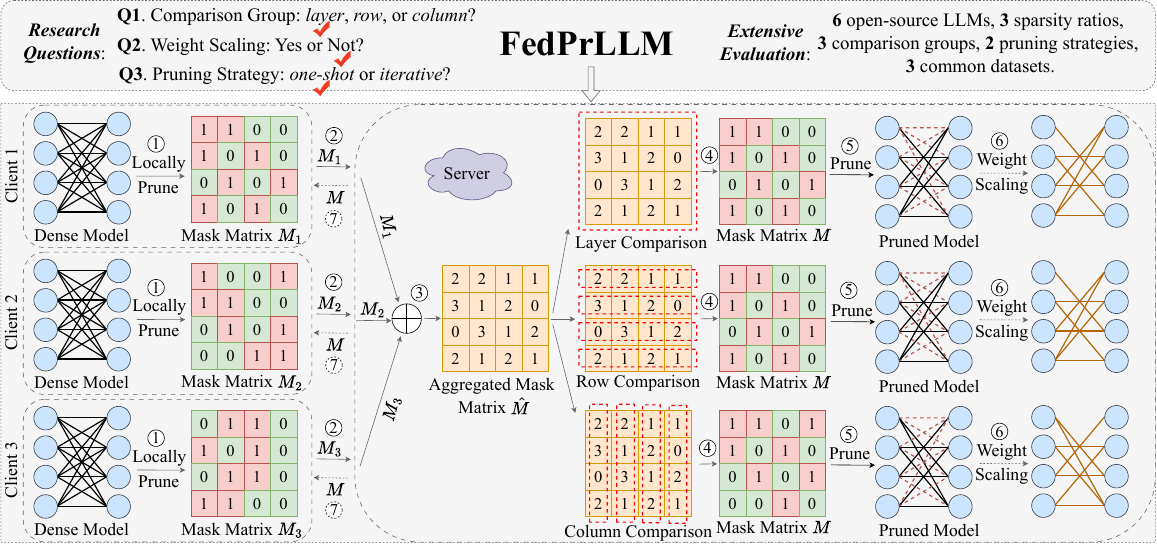}
    \vspace{-0.5cm}
    \caption{\textit{Top)}. Research questions alongside the corresponding findings and experimental scenarios.
    \textit{Bottom)}. The {FedPrLLM} framework. \textcircled{\scriptsize{1}}
    Each client calculates a pruning mask matrix $\mathbf{M}_i$ using its calibration dataset $\mathcal{D}_i$. \textcircled{\scriptsize{2}} Clients send the mask matrices $\mathbf{M}_i$ to the server. \textcircled{\scriptsize{3}} The server aggregates these mask matrices $\mathbf{M}_i$ to obtain an aggregated mask matrix $\mathbf{\hat{M}} = \sum_{i=1}^{m}\mathbf{M}_i$. \textcircled{\scriptsize{4}} Top-k values are selected from the aggregated mask matrix $\mathbf{\hat{W}}$ to derive the final mask matrix  $\mathbf{M}$. \textcircled{\scriptsize{5}} Prune the global model $\mathbf{W}$ using the mask matrix $\mathbf{M}$ as follows: $\mathbf{\hat{W}} = \mathbf{W} \odot (1 - \mathbf{M})$, where $\odot$ denotes element-wise multiplication. \textcircled{\scriptsize{6}} Scale the model weights of the retained parameters using the aggregated mask matrix $\mathbf{\hat{M}}$ as follows: $\mathbf{\hat{W}} \odot \frac{(m - \mathbf{\hat{M}})}{m}$ (if needed). \textcircled{\scriptsize{7}} The server broadcasts the mask matrix $\mathbf{M}$ to each client (for iterative pruning). The dashed arrow indicates that this operation is optional; step \textcircled{\scriptsize{6}} is used for weight scaling, while \textcircled{\scriptsize{7}} is used for iterative pruning. Note that this visualization is primarily for one-shot pruning, which requires only one communication round. For iterative pruning, multiple communication rounds will occur between steps \textcircled{\scriptsize{2}} and \textcircled{\scriptsize{7}}, and the layer index is omitted here.
    }
    \label{fig:framework}
    \vspace{-0.4cm}
\end{figure}

To address these challenges, we formalize the first systematic study on federated LLM pruning and empirically evaluate three core design choices through a unified \textbf{FedPrLLM} framework (Figure \ref{fig:framework}):
\vspace{-0.1cm}

\begin{center}
\raggedright

    \hspace{2em}
    \textbf{Q1.} \textit{Comparison Group}: Which comparison group is more effective: \textit{layer}, \textit{row}, or \textit{column}?

    \hspace{2em}
    \textbf{Q2.} \textit{Weight Scaling}: Should we scale the model weights of the retained parameters?

    \hspace{2em}
    \textbf{Q3.} \textit{Pruning Strategy}: Does iterative pruning outperform one-shot pruning?
\end{center}

We dedicated thousands of GPU hours to benchmark federated pruning for LLMs, conducting extensive experiments across \textbf{6} open-source LLMs, \textbf{3} sparsity ratios, \textbf{3} comparison groups, \textbf{2} pruning strategies on \textbf{3} common datasets.
From these efforts, we have developed a practical list of key insights for federated pruning of LLMs:
\vspace{-0.1cm}
{
\renewcommand{\theenumi}{\textbf{\arabic{enumi}).}}
\renewcommand{\labelenumi}{\theenumi}
\begin{enumerate}
    \item \textbf{Layer comparison is simple yet effective.} Among the three comparison groups—\textit{layer}, \textit{row}, and \textit{column comparisons}—layer comparison stands out as the simplest and most effective method, regardless of the local pruning method's comparison group.
    \item \textbf{Scaling weights performs worse than expected.} Though the FL aggregated mask matrix, which reveals how strongly each parameter is disfavored across clients, could be used to scale the retained parameters for continuous weight adjustment, its performance is inferior to that of not scaling them.
    \item \textbf{Iterative pruning offers no benefit.} While iterative pruning allows for prompt refinement of the local model, it incurs significant communication overhead, and its performance is comparable to that of one-shot pruning, offering no additional advantages.

\end{enumerate}
}

We hope our findings will help guide future efforts in federated pruning for LLMs and inform best practices for deploying LLMs under federated scenarios in real-world applications. We summarize our contributions as follows:

\begin{itemize}
    \item We introduce \textbf{FedPrLLM}, a comprehensive federated pruning framework designed for the privacy-preserving compression of LLMs, which incorporates various possibilities for integrating FL with LLM pruning. 
    \item We conduct an extensive evaluation of FedPrLLM, providing practical insights into effective federated pruning techniques for LLMs, based on thousands of GPU hours invested in multiple open-source LLMs, various sparsity ratios, comparison groups, and datasets.
    \item We identify that layer comparison is simple yet effective, scaling weights offers no benefits and may worsen performance, and that one-shot pruning is as effective as iterative pruning while reducing communication costs.
\end{itemize}

\vspace{-0.2cm}
\section{Preliminaries}
\label{sec:preliminaries}

\subsection{LLM Pruning}

LLM pruning can be broadly classified into \textit{structured pruning} \cite{ma2023llm,li2023losparse,ashkboos2024slicegpt,xia2024sheared,gromov2025the} and \textit{unstructured pruning} \cite{frantar2023sparsegpt,sun2024a,zhang2024plug,yin2024outlier,xu2024besa}, and in this work, we focus on the latter. Unstructured pruning involves removing individual weights from the model's weight matrices based on certain criteria while maintaining its performance as much as possible \cite{frantar2023sparsegpt,sun2024a,zhang2024plug,yin2024outlier,xu2024besa}. It is usually achieved by minimizing the discrepancy square error between the dense and pruned model \textit{layer-by-layer} recursively. Specifically, for an uncompressed linear layer with weights $\mathbf{W}_l \in \mathbb{R}^{d \times r}$, the objective for unstructured pruning can usually be formulated as:
\begin{equation}
    \mathop{\arg\min}_{\mathbf{M}_l} \|\mathbf{W}_l \mathbf{X}_l - (\mathbf{W}_l \odot	 (1-\mathbf{M}_l))\mathbf{X}_l \|_2^2 \quad \text{s.t.} \quad \|\mathbf{M}_l\|_0 \geq k,
\end{equation}
where $\mathbf{X}_l$ is the input to $l$-th linear layer (also referred to as calibration data),  $\mathbf{M}_l \in \{0, 1\}^{d \times r}$ is the pruning mask matrix we aim to derive, $\odot$ denotes element-wise multiplication, $\| \cdot \|_0$ is the $l_0$-norm (e.g., the number of non-zero elements), and $k$ represents the number of pruned weights determined by the pruning ratio. 

The differences between previous pruning methods primarily lie in the design of the pruning metrics and the comparison groups used to derive the pruning mask matrix \cite{frantar2023sparsegpt,sun2024a,zhang2024plug}. Pruning metrics refer to how the importance of each model weight is identified, while comparison groups denote the selection of groups for comparing these weights, including \textit{layer comparison}, \textit{row comparison}, and \textit{column comparison}. For example, SparseGPT \cite{frantar2023sparsegpt} utilizes the Hessian Matrix inverse, i.e., $\left[\frac{|\mathbf{W}|^2}{\rm{diag} \left(\left(\mathbf{X}^T\mathbf{X} + \lambda \mathbf{I}\right)^{-1}\right)}\right]_{ij}$, as the pruning metric, employing layer comparison to determine the pruning mask matrix for pruning, along with subsequent weight scaling. Wanda \cite{sun2024a} adopts the magnitudes of model weights multiplied by the corresponding input activations, i.e., $|\mathbf{W}_{ij}| \cdot \|\mathbf{X}_j\|_2$, as the pruning metric and chooses row comparison. RIA \cite{zhang2024plug} integrates relative importance within Wanda, resulting in a new pruning metric, i.e., $\left(\frac{|\mathbf{W}_{ij}|}{\sum |\mathbf{W}_{{*j}|}} + \frac{|\mathbf{W}_{ij}|}{\sum |\mathbf{W}_{{i*}|}}\right) \cdot \left( \|\mathbf{X}_j\|_2\right)^{0.5}$, while utilizing layer comparison.

\subsection{Federated Learning}

Federated Learning (FL) \citep{mcmahan2017communication,zhang2024flhetbench,zeng2024tackling,guo2025new,guo2025exploring,guo2025selective} is a decentralized approach where multiple clients collaboratively train a model while keeping their data localized. The objective is to minimize a global loss function:
\begin{equation}
\mathcal{L}(\Theta) = \frac{1}{m} \sum_{i=1}^m \ell_i(\Theta),
\end{equation}
where $\Theta$ denotes the model weight, $m$ is the total number of clients, 
and $\ell_i(\Theta)$ is the local loss function for client $i$ . 

In FL, each client trains its model using its local dataset, aiming to find:
$
\hat{\Theta}_i = \arg\min_{\Theta} \ell_i(\Theta).
$
After local training, clients send their model updates $\Delta \Theta_i$ to a central server, which aggregates these updates to update the global model:
$
\Theta \leftarrow \Theta + \frac{1}{m} \sum_{i=1}^m \Delta \Theta_i.
$
The local training and server aggregation process typically requires multiple communication rounds until the model converges or reaches the predefined maximum number of rounds.


\section{Federated Pruning for LLMs}
\label{sec:method}

\subsection{Problem Formulation}

In the federated pruning scenario for LLMs, multiple clients aim to collaboratively prune an LLM while ensuring that their local calibration data remains private. Formally, let $\mathbf{W}$ represent the model parameters of the LLM that we aim to prune. Each client $i$ possesses a private calibration dataset denoted as $\mathcal{D}_i$, which is used for calculating the pruning mask matrices during the local pruning process. 
These mask matrices are then shared with the server to prune the LLM.

\subsection{FedPrLLM}

In this section, we first introduce the overall workflow of the comprehensive \textbf{FedPrLLM} framework, as illustrated at the bottom of Figure \ref{fig:framework}, and then discuss the various possibilities within it. Specifically, during local pruning, each client calculates a pruning mask matrix $\mathbf{M}_i \in \{0, 1\}^{|\mathbf{W}_i|}$ using its calibration dataset $\mathcal{D}_i$ (step \textcircled{\scriptsize{1}}). This mask matrix determines which weights are pruned ($\mathbf{M}_{ij} = 1$) and which are retained ($\mathbf{M}_{ij} = 0$). The decision on which weights to prune or retain is based on an importance criterion derived from the calibration data, such as the magnitudes of model weights multiplied by the corresponding input activations used in Wanda \cite{sun2024a}.

After calculating the pruning mask matrix, each client $i$ shares only the mask matrix $\mathbf{M}_i$ with the central server (step \textcircled{\scriptsize{2}}). This approach ensures that no local model parameters or private calibration data are transmitted, thereby minimizing communication overhead and preserving data privacy. Upon receiving the pruning mask matrices $\mathbf{M}_i$ from all clients, the server sums them to obtain an aggregated mask matrix $\mathbf{\hat{M}} = \sum_{i=1}^{m}\mathbf{M}_i$ (step \textcircled{\scriptsize{3}}) and then selects the top-k values to create the final mask matrix $\mathbf{M}$ (step \textcircled{\scriptsize{4}}) for pruning the global model (step \textcircled{\scriptsize{5}}).
In the following, we will discuss various possibilities within the FedPrLLM framework, including different comparison groups, the decision to perform weight scaling, and the choice between one-shot and iterative pruning.

\subsubsection{Comparison Group}
\label{sec:compare_group}

When selecting the top-k values from the aggregated mask matrix $\mathbf{\hat{M}}$ to derive the final pruning mask matrix $\mathbf{M}$, three comparison groups can be considered (step \textcircled{\scriptsize{4}}): \textit{layer comparison}, \textit{row comparison}, and \textit{column comparison}. In layer comparison, the comparison group consists of all elements within a layer, allowing us to choose the top-k values across the entire layer. Conversely, in row (or column) comparison, the comparison group is defined by each individual row (or column), enabling the selection of the top-k values within each respective row (or column). The visualization of these comparison groups is shown in Figure \ref{fig:framework}. Thus, given that multiple comparison groups could be chosen, \textit{which comparison group is more effective for federated pruning of LLMs}?

\subsubsection{Weight Scaling}

After obtaining the final mask matrix $\mathbf{M}$, it can be used to effectively prune the dense model $\mathbf{W}$ using $\mathbf{W} \odot (1 - \mathbf{M})$, where $\odot$ denotes element-wise multiplication (step \textcircled{\scriptsize{5}}). This operation removes the weights corresponding to the masked parameters (i.e., $\mathbf{M}_{ij}=1$), resulting in a sparser model $\mathbf{\hat{W}}$. 

Then, beyond merely determining which parameters to prune via majority voting (i.e., selecting top-k values), the aggregated mask matrix $\mathbf{\hat{M}}$ reveals a critical hidden signal: how strongly each parameter is disfavored across clients. Consider two surviving parameters - one narrowly retained (pruned by 10/100 clients) and another unanimously preserved (pruned by 0/100 clients). Traditional pruning treats both equally, maintaining their original magnitudes despite their differing consensus levels. However, this ignores a critical insight: the former parameter, though retained, exhibits weaker consensus across clients. To this end, the aggregated mask matrix $\mathbf{\hat{M}}$ could be further applied to scale down the retained parameters using the formula $\mathbf{\hat{W}} \odot \frac{(m - \mathbf{\hat{M}})}{m}$ (step \textcircled{\scriptsize{6}}, if needed). 
This approach corresponds to locally pruning the model and then sharing the pruned model with the server, which aggregates them using the FedAvg algorithm \cite{mcmahan2017communication}. However, \textit{will the weight scaling improve the performance of federated pruning for LLMs}?

\subsubsection{One-shot vs. Iterative Pruning}

Since LLMs are usually pruned \textit{layer-by-layer} recursively \cite{frantar2023sparsegpt,sun2024a,zhang2024plug}, federated pruning for LLMs can be naturally categorized into two types: \textit{one-shot pruning} and \textit{iterative pruning}. 
In one-shot pruning, each client calculates the pruning mask matrices for all layers and then sends them to the server, resulting in only one communication round.
In contrast, iterative pruning involves sending the pruning mask matrices to the server layer by layer. Specifically, after calculating the pruning mask matrix for one layer, it is uploaded to the server for aggregation. The server then combines these matrices into a global mask matrix for pruning the model at that layer and broadcasts the global mask matrix back to each client for local pruning of that layer (step \textcircled{\scriptsize{7}}, the layer index is omitted here). This process is carried out layer by layer and involves multiple communication rounds, resulting in higher communication costs compared to one-shot pruning. Therefore, given the significant communication costs associated with iterative pruning, \textit{will iterative pruning outperform one-shot pruning}?

One-shot and iterative pruning differ because, when calculating the pruning mask matrix for layer $l+1$ locally, the calibration data $\mathbf{X}_{l+1}$ is derived from the output of layer $l$, which has already been pruned. Since the weights of the local pruned model for layer $l$ vary between using $\mathbf{M}_i$ (one-shot pruning) and $\mathbf{M}$ (iterative pruning), this leads to different outputs for layer $l$ and, consequently, varying calibration data $\mathbf{X}_{l+1}$, resulting in distinct pruning mask matrices for layer $l+1$.

\section{Experiments}
\label{sec:exp}

Our experiments are designed to answer the following research questions that are important for the practical pruning of LLMs under a federated scenario.

\begin{itemize}
    \item \textbf{Q1.} Which comparison group is more effective: \textit{layer}, \textit{row}, or \textit{column}?
    \item \textbf{Q2.} Should we scale the model weights of the retained parameters?
    \item \textbf{Q3.} Does iterative pruning outperform one-shot pruning?
\end{itemize}

\subsection{Experimental Setup} 
\label{sec:exp_details}

We implement FedPrLLM in PyTorch \cite{NEURIPS2019_bdbca288} and use the Hugging Face Transformers library \cite{wolf2019huggingface} to handle models and datasets. We evaluate the FedPrLLM on the three most widely adopted LLM model families: LLaMA 7B/13B/30B \cite{touvron2023llama}, LLaMA-2 7B/13B \cite{touvron2023llama2} and LLaMA-3 8B \cite{meta2024introducing}. For each model under consideration, we focus on pruning the linear layers (skipping the first embedding layer and the final classification head), which account for around 99\% of the total LLM parameters. We employ unstructured sparsity and impose a uniform sparsity ratio for all linear layers. 

For the calibration data, following \cite{frantar2023sparsegpt,sun2024a,xu2024besa,zhang2024plug}, we use 128 samples from the C4 dataset \cite{raffel2020exploring}, with each sample containing 2048 tokens. For FedPrLLM, we set the number of clients to 64, resulting in each client having only 2 calibration samples. For each client, we adopt Wanda \cite{sun2024a} to perform local pruning and calculate the pruning mask matrix.

Apart from the proposed FedPrLLM framework, we further implement two baselines for comparison: (1) \textbf{Local-only}, where each client prunes the model locally using its private calibration data, and (2) \textbf{Centralized}, where the server prunes the model with all calibration data, which could be considered as an upper bound for the pruning performance under FL setting. 

Following previous works on LLM compression \cite{frantar2023sparsegpt,xu2024besa,zhang2024plug}, we measure the performance of pruned models in language modeling and evaluate their perplexity on the held-out WikiText2 \cite{merity2017pointer} validation set, C4 \cite{raffel2020exploring} validation data, and PTB \cite{marcus1994penn}. 

All experiments are conducted on NVIDIA L40S GPUs.

\subsection{Main Results}

To answer the research questions above, we conducted extensive experiments to evaluate FedPrLLM along with two baselines across \textbf{6} open-source LLMs, \textbf{3} sparsity ratios, \textbf{3} comparison groups, \textbf{2} pruning strategies on \textbf{3} common datasets. The experimental results for the 50\% sparsity ratio are shown in Tables \ref{tab:main_results_wiki}, \ref{tab:main_results_c4}, and \ref{tab:main_results_ptb}, while results for higher sparsity ratios (e.g., 60\% and 70\%) are shown in Tables \ref{app_tab:main_results_wiki}, \ref{app_tab:main_results_c4}, and \ref{app_tab:main_results_ptb} in Appendix.

\begin{table} [t]
\centering
\caption{WikiText2 perplexity of pruned LLMs under 50\% sparsity ratio.}
\label{tab:main_results_wiki}
    \resizebox{\linewidth}{!}{
    \begin{tabular}{l lcc ccc cc c}
    \toprule
        & Compar. & Prune & Weight & \multicolumn{3}{c}{LLaMA} & \multicolumn{2}{c}{LLaMA-2} & LLaMA-3 \\
        \cmidrule(lr){5-7} \cmidrule(lr){8-9} \cmidrule(lr){10-10} 
        Method & Group & Stra. & Scaling & 7B & 13B & 30B & 7B & 13B & 8B \\
        \midrule
        Dense & - & - & - & 5.67 & 5.09 & 4.10 & 5.11 & 4.57 & 7.46 \\
        \midrule
        Centralized & - & - & - & 7.25 & 6.15 & 5.24 & 6.46 & 5.58 & 11.00 \\
        Local-only & - & - & - & 7.44 & 6.33 & 5.34 & 6.63 & 5.72 & 11.39 \\
        \cmidrule(lr){1-10}
        \multirow{12}{*}{FedPrLLM} & Layer & One-shot & \ding{55} & 7.32 & \textbf{6.19} & \textbf{5.24} & \textbf{6.48} & \textbf{5.61} & \textbf{11.02} \\ 
        & Row & One-shot & \ding{55} & \textbf{7.30} & 6.20 & 5.25 & \textbf{6.48} & \textbf{5.61} & \textbf{11.02} \\
        & Column & One-shot & \ding{55} & 1524.28 & 9282.09 & 501.88 & 20528.41 & 5309.48 & 311468.53 \\
        & Layer & Iterative & \ding{55} & \textbf{7.30} & \textbf{6.19} & \textbf{5.24} & \textbf{6.48} & 5.62 & 11.12 \\ 
        & Row & Iterative & \ding{55} & \textbf{7.30} & 6.20 & \textbf{5.24} & \textbf{6.48} & \textbf{5.61} & 11.11 \\
        & Column & Iterative & \ding{55} & 1822.89 & 6884.15 & 996.57 & 77245.84 & 5430.81 & 189134.78 \\
        & Layer & One-shot & \checkmark & 7.48 & 6.36 & 5.35 & 6.67 & 5.75 & 11.75 \\ 
        & Row & One-shot & \checkmark & 7.47 & 6.36 & 5.35 & 6.67 & 5.75 & 11.75 \\
        & Column & One-shot & \checkmark & 1708.41 & 10819.42 & 824.50 & 18084.02 & 5914.91 & 276031.34 \\
        & Layer & Iterative & \checkmark & 7.46 & 6.35 & 5.34 & 6.67 & 5.75 & 11.86 \\ 
        & Row & Iterative & \checkmark & 7.46 & 6.35 & 5.34 & 6.67 & 5.74 & 11.87 \\
        & Column & Iterative & \checkmark & 1985.40 & 6692.91 & 939.62 & 66911.49 & 5268.71 & 41996.95 \\
    \bottomrule
    \end{tabular}
    }
\vspace{-0.2cm}
\end{table}

\begin{table} [t]
\centering
\caption{C4 perplexity of pruned LLMs under 50\% sparsity ratio.}
\label{tab:main_results_c4}
    \resizebox{\linewidth}{!}{
    \begin{tabular}{l lcc ccc cc c}
    \toprule
        & Compar. & Prune & Weight & \multicolumn{3}{c}{LLaMA} & \multicolumn{2}{c}{LLaMA-2} & LLaMA-3 \\
        \cmidrule(lr){5-7} \cmidrule(lr){8-9} \cmidrule(lr){10-10} 
        Method & Group & Stra. & Scaling & 7B & 13B & 30B & 7B & 13B & 8B \\
        \midrule
        Dense & - & - & - & 7.34 & 6.79 & 6.12 & 7.03 & 6.51 & 12.34 \\
        \midrule
        Centralized & - & - & - & 9.34 & 8.14 & 7.28 & 8.94 & 8.03 & 18.38  \\
        Local-only & - & - & - & 9.59 & 8.37 & 7.52 & 9.16 & 8.31 & 18.92 \\
        \cmidrule(lr){1-10}
        \multirow{12}{*}{FedPrLLM} & Layer & One-shot & \ding{55} & \textbf{9.43} & \textbf{8.22} & \textbf{7.39} & \textbf{9.01} & \textbf{8.18} & \textbf{18.32} \\ 
        & Row & One-shot & \ding{55} & \textbf{9.43} & \textbf{8.22} & \textbf{7.39} & \textbf{9.01} & 8.19 & \textbf{18.32} \\
        & Column & One-shot & \ding{55} & 893.05 & 10616.94 & 512.27 & 9631.37 & 5075.92 & 200257.70 \\
        & Layer & Iterative & \ding{55} & 9.44 & \textbf{8.22} & \textbf{7.39} & \textbf{9.01} & 8.19 & 18.43 \\ 
        & Row & Iterative & \ding{55} & 9.44 & \textbf{8.22} & \textbf{7.39} & 9.02 & \textbf{8.18} & 18.38 \\
        & Column & Iterative & \ding{55} & 1050.26 & 8567.66 & 779.01 & 11658.80 & 4804.46 & 112192.42 \\
        & Layer & One-shot & \checkmark & 9.64 & 8.40 & 7.57 & 9.21 & 8.39 & 19.45 \\ 
        & Row & One-shot & \checkmark & 9.64 & 8.40 & 7.57 & 9.21 & 8.39 & 19.45 \\
        & Column & One-shot & \checkmark & 887.34 & 13744.66 & 895.18 & 11440.51 & 5189.73 & 90476.94 \\
        & Layer & Iterative & \checkmark & 9.64 & 8.41 & 7.57 & 9.22 & 8.39 & 19.58 \\ 
        & Row & Iterative & \checkmark & 9.65 & 8.41 & 7.57 & 9.22 & 8.39 & 19.60 \\
        & Column & Iterative & \checkmark & 1242.31 & 6860.69 & 724.28 & 10355.87 & 4657.88 & 44469.52 \\
    \bottomrule
    \end{tabular}
    }
\end{table}

\begin{table} [t]
\centering
\caption{PTB perplexity of pruned LLMs under 50\% sparsity ratio.}
\label{tab:main_results_ptb}
    \resizebox{\linewidth}{!}{
    \begin{tabular}{l lcc ccc cc c}
    \toprule
        & Compar. & Prune & Weight & \multicolumn{3}{c}{LLaMA} & \multicolumn{2}{c}{LLaMA-2} & LLaMA-3 \\
        \cmidrule(lr){5-7} \cmidrule(lr){8-9} \cmidrule(lr){10-10} 
        Method & Group & Stra. & Scaling & 7B & 13B & 30B & 7B & 13B & 8B \\
        \midrule
        Dense & - & - & - & 41.15 & 28.09 & 23.51 & 50.20 & 56.51 & 13.30 \\
        \midrule
        Centralized & - & - & - & 80.12 & 36.41 & 26.64 & 96.99 & 86.83 & 20.69 \\
        Local-only & - & - & - & 86.25 & 37.57 & 27.13 & 108.66 & 91.92 & 21.43 \\
        \cmidrule(lr){1-10}
        \multirow{12}{*}{FedPrLLM} & Layer & One-shot & \ding{55} & \textbf{80.31} & 36.57 & 26.69 & 102.71 & \textbf{88.26} & 20.56 \\ 
        & Row & One-shot & \ding{55} & 80.71 & 36.61 & \textbf{26.64} & \textbf{101.85} & 88.31 & \textbf{20.55} \\
        & Column & One-shot & \ding{55} & 4463.92 & 22138.56 & 713.56 & 14256.86 & 7392.64 & 407313.84 \\
        & Layer & Iterative & \ding{55} & 81.22 & \textbf{36.54} & 26.68 & 102.72 & 88.38 & \textbf{20.55} \\ 
        & Row & Iterative & \ding{55} & 81.26 & 36.55 & \textbf{26.64} & 103.66 & 88.94 & 20.60 \\
        & Column & Iterative & \ding{55} & 4061.96 & 17610.52 & 1158.75 & 13401.63 & 6941.72 & 168643.04 \\
        & Layer & One-shot & \checkmark & 87.97 & 37.70 & 27.27 & 112.52 & 92.90 & 22.21 \\ 
        & Row & One-shot & \checkmark & 88.35 & 37.72 & 27.25 & 112.17 & 93.07 & 22.21 \\
        & Column & One-shot & \checkmark & 4557.48 & 29140.28 & 982.59 & 12021.08 & 7801.23 & 264723.12 \\
        & Layer & Iterative & \checkmark & 87.28 & 37.69 & 27.27 & 112.95 & 92.58 & 22.39 \\ 
        & Row & Iterative & \checkmark & 87.61 & 37.60 & 27.27 & 113.32 & 92.61 & 22.41 \\
        & Column & Iterative & \checkmark & 6929.83 & 15189.83 & 1178.40 & 10208.03 & 5220.64 & 39172.53 \\
    \bottomrule
    \end{tabular}
    }
\vspace{-0.2cm}
\end{table}

\subsubsection{Which Comparison Group is More Effective?}

As discussed above, various comparison groups can be used to select top-k values from the aggregated mask matrix to derive the final mask matrix for pruning the global model, including \textit{layer comparison}, \textit{row comparison}, and \textit{column comparison}. Thus, which comparison group is the most effective?

According to the results in Tables \ref{tab:main_results_wiki}, \ref{tab:main_results_c4}, and \ref{tab:main_results_ptb}, we observe that layer comparison and row comparison achieve comparable performance, both significantly surpassing column comparison. To investigate why column comparison performs much worse than the others, we noted that the local pruning method we used (i.e., Wanda \cite{sun2024a}) adopts row comparison, meaning the local pruning mask matrix $\mathbf{M}_i$ derived from each client is based on row comparison. We hypothesize that this is the reason for the poorer performance of column comparison, as the comparison group used in FedPrLLM conflicts with that of the local pruning method.

To validate this, we further change the comparison group in the local pruning method (i.e., Wanda \cite{sun2024a}) to layer comparison and column comparison to evaluate the performance of the FedPrLLM framework with one-shot pruning and no weight scaling. The results on WikiText2 are shown in Table \ref{tab:evaluate_compar_group_wiki}, while results for other datasets are presented in Table \ref{app_tab:evaluate_compar_group} in Appendix. From these results, we see that when the comparison group in the local pruning method (i.e., Wanda \cite{sun2024a}) is changed to layer comparison, only the layer comparison used in FedPrLLM performs well, while row comparison performs poorly and column comparison performs even worse. Similarly, when the local pruning method's comparison group is changed to column comparison, only the layer and column comparisons perform normally, while row comparison performance is poor. Note that when the comparison group in the local pruning method (i.e., Wanda \cite{sun2024a}) is changed to column comparison, it degrades to the magnitude-based pruning method, rendering the performance irrelevant to calibration samples, which results in the performance of Centralized and Local-only being the same \cite{sun2024a}. These results demonstrate our hypothesis that the conflict between the local and server comparison groups leads to worse performance, while the layer comparison used in FerPrLLM consistently achieves good results, regardless of the comparison group used for the local pruning method. Therefore, we conclude that:

\begin{takeaway}
    Layer comparison is simple yet effective.
\end{takeaway}

\begin{table} [t]
\centering
\caption{WikiText2 perplexity of pruned LLMs under 50\% sparsity ratio when changing the comparison group for the local pruning method. FedPrLLM adopts one-shot pruning and no weight scaling.}
\label{tab:evaluate_compar_group_wiki}
    \resizebox{\linewidth}{!}{
    \begin{tabular}{ccl ccc cc c}
    \toprule
    Local Compar. & & Compar. & \multicolumn{3}{c}{LLaMA} & \multicolumn{2}{c}{LLaMA-2} & LLaMA-3 \\
        \cmidrule(lr){4-6} \cmidrule(lr){7-8} \cmidrule(lr){9-9} 
    Group & Method & Group & 7B & 13B & 30B & 7B & 13B & 8B \\
        \midrule
        \multirow{5}{*}{Layer} & Centralized & - & 7.94 & 6.57 & 5.47 & 7.38 & 5.92 & 12.04 \\ 
        & Local-only & - & 8.16 & 6.74 & 5.58 & 7.56 & 6.06 & 12.43 \\
        \cmidrule(lr){2-9}
        & \multirow{3}{*}{FedPrLLM} & Layer & \textbf{7.98} & \textbf{6.60} & \textbf{5.48} & \textbf{7.38} & \textbf{5.95} & \textbf{12.09} \\
        &  & Row & 31.85 & 10.08 & 11.33 & 39.07 & 124.08 & 17.51 \\
        &  & Column & 1749.59 & 10183.32 & 541.62 & 25258.16 & 5503.91 & 336255.96 \\
        \midrule
        \multirow{5}{*}{Column} & Centralized & - & 8.86 & 7.68 & 5.67 & 10.41 & 6.38 & 83.67 \\ 
        & Local-only & - & 8.86 & 7.68 & 5.67 & 10.41 & 6.38 & 83.67 \\
        \cmidrule(lr){2-9}
        & \multirow{3}{*}{FedPrLLM} & Layer & \textbf{8.86} & \textbf{7.68} & \textbf{5.67} & \textbf{10.41} & \textbf{6.38} & \textbf{83.67} \\
        &  & Row & 138.54 & 100.80 & 49.17 & 764.32 & 2580.88 & 400.95 \\
        &  & Column & \textbf{8.86} & \textbf{7.68} & \textbf{5.67} & \textbf{10.41} & \textbf{6.38} & \textbf{83.67} \\
    \bottomrule
    \end{tabular}
    }
\end{table}

\subsubsection{Should We Scale the Model Weights of the Retained Parameters?}

The aggregated mask matrix $\mathbf{\hat{M}}$ indicates the number of clients that wish to prune a parameter, which allows it to be used for scaling the model weights of the retained parameters to $\frac{(m - \mathbf{\hat{M}})}{m}$. This approach corresponds to locally pruning the model and then sharing the pruned model with the server, which aggregates them using the FedAvg algorithm \cite{mcmahan2017communication}. 
However, will weight scaling be beneficial for the federated pruning of LLMs?

From the results in Tables \ref{tab:main_results_wiki}, \ref{tab:main_results_c4}, and \ref{tab:main_results_ptb}, we observe that the performance with weight scaling is worse than that without weight scaling across all comparison groups and pruning strategies. It indicates that scaling weights offers no benefit and may even worsen performance. This may be due to the fact that locally pruned models do not perform well, and applying the FedAvg algorithm \cite{mcmahan2017communication} to aggregate these pruned model weights leads to subpar performance. 
Therefore, we conclude that:

\begin{takeaway}
    Scaling weights performs worse than expected.
\end{takeaway}

\subsubsection{Does Iterative Pruning Outperform One-shot Pruning?}

Since LLMs are usually pruned \textit{layer-by-layer} recursively \cite{frantar2023sparsegpt,sun2024a,zhang2024plug}, federated pruning for LLMs can be naturally categorized into two types: \textit{one-shot pruning} and \textit{iterative pruning}. 
Given the significant communication costs associated with iterative pruning, will it outperform one-shot pruning?

The comparison results are provided in Tables \ref{tab:main_results_wiki}, \ref{tab:main_results_c4}, and \ref{tab:main_results_ptb}. These results indicate that the performance of iterative pruning and one-shot pruning is comparable, regardless of the comparison groups and pruning strategies. However, since iterative pruning introduces significant communication costs without any performance improvement, we conclude that:

\begin{takeaway}
    Iterative pruning offers no benefit.
\end{takeaway}

\subsection{Sensitivity Analysis}

In this section, we conduct sensitivity analyses on the number of clients and calibration samples in FedPrLLM to better understand its effectiveness in pruning LLMs within a federated scenario. We use FedPrLLM, which employs layer comparison, one-shot pruning, and no weight scaling, to conduct the analysis under a 50\% sparsity ratio.

\subsubsection{Impact of Client Numbers} 

\begin{figure}[ht]
\vspace{-0.2cm}
\centering
\subfigure[WikiText2] { \label{fig:ablation_NC_wiki}
\includegraphics[width=0.32\columnwidth]{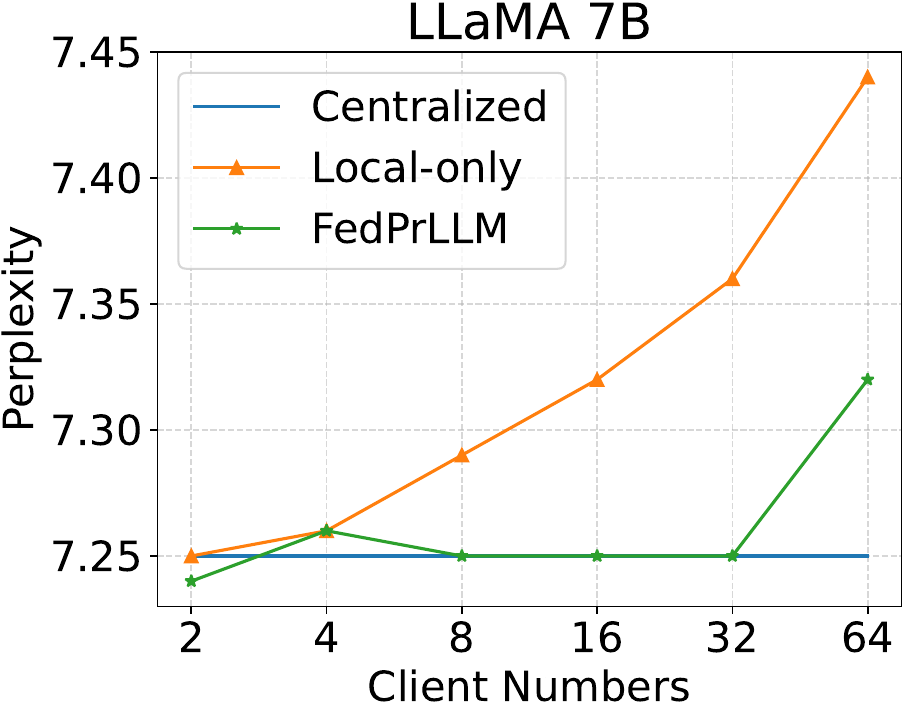}}   
\hfill
\subfigure[C4] { \label{fig:ablation_NC_c4} 
\includegraphics[width=0.32\columnwidth]{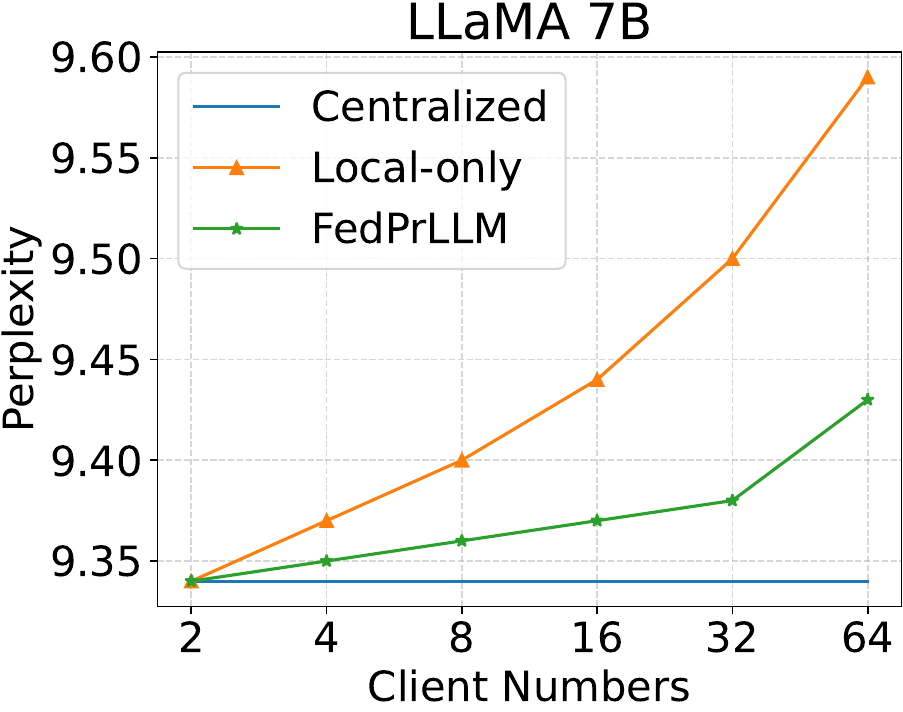}} 
\subfigure[PTB] { \label{fig:ablation_NC_ptb} 
\includegraphics[width=0.32\columnwidth]{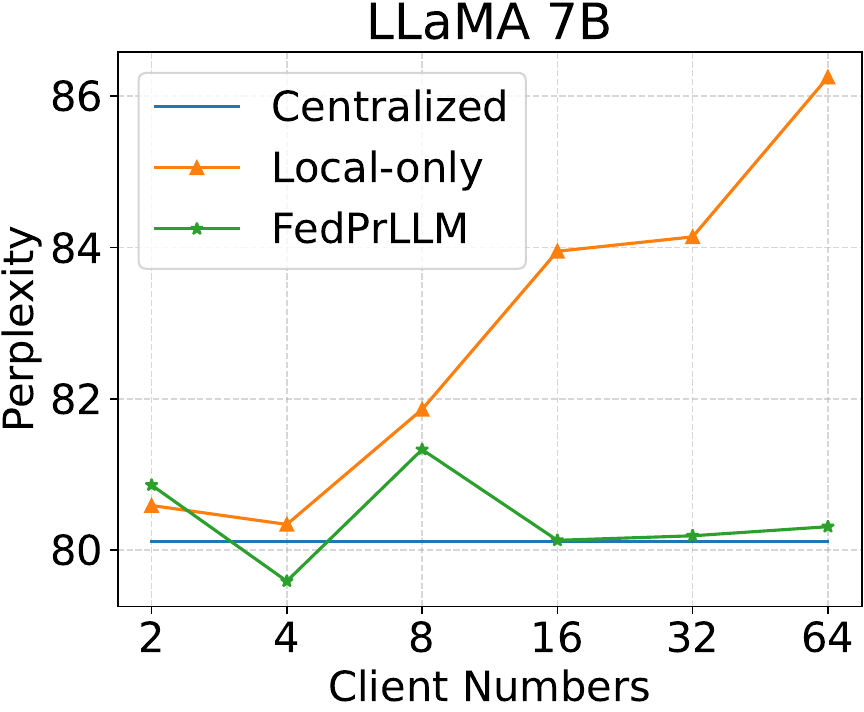}} 
\vspace{-0.1cm}
\caption{The effect of different client numbers on federated pruning LLMs.}
\label{fig:ablation_NC}
\vspace{-0.2cm}
\end{figure}

It is worth noting that the number of clients influences the performance of FL algorithms \cite{guo2025new,guo2025selective}. In this section, we investigate the effect of client numbers on the federated pruning of LLMs. We use a total of 128 calibration samples and vary the number of clients from 64 to 2, resulting in an increase in the calibration samples allocated to each client. Specifically, when the number of clients is 64, each client has only 2 calibration samples; when the number of clients is reduced to 2, each client has 64 calibration samples. The experimental results are shown in Figure \ref{fig:ablation_NC}. From this figure, we observe that FedPrLLM consistently outperforms Local-only pruning across various numbers of clients, demonstrating the effectiveness of the federated pruning algorithm.

\subsubsection{Impact of the Number of Calibration Samples}

\begin{figure}[ht]
\vspace{-0.2cm}
\centering
\subfigure[WikiText2] { \label{fig:ablation_NS_wiki}
\includegraphics[width=0.32\columnwidth]{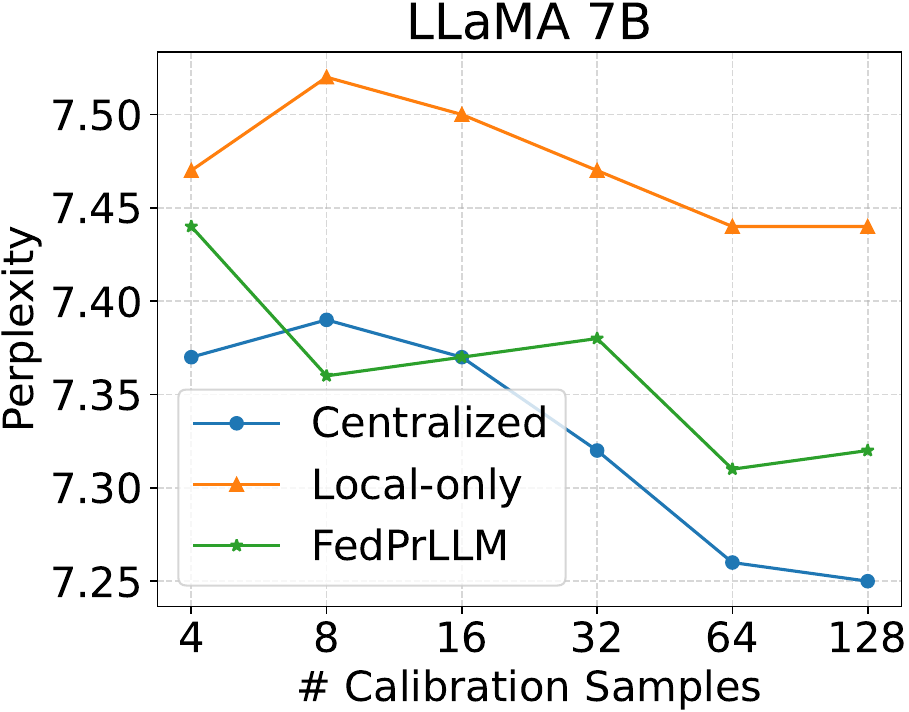}}   
\hfill
\subfigure[C4] { \label{fig:ablation_NS_c4} 
\includegraphics[width=0.32\columnwidth]{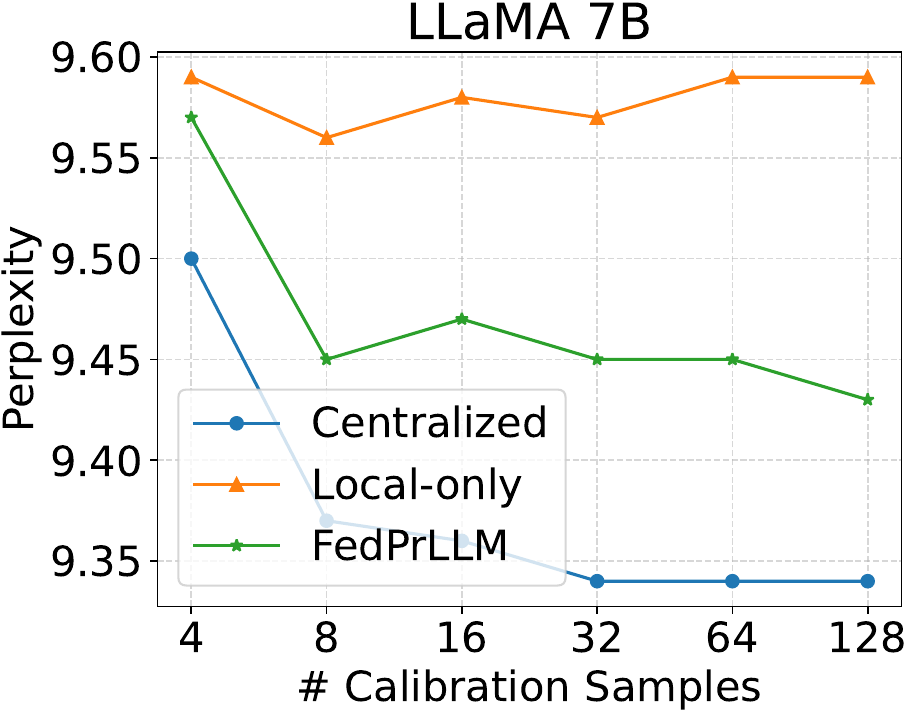}} 
\subfigure[PTB] { \label{fig:ablation_NS_ptb} 
\includegraphics[width=0.32\columnwidth]{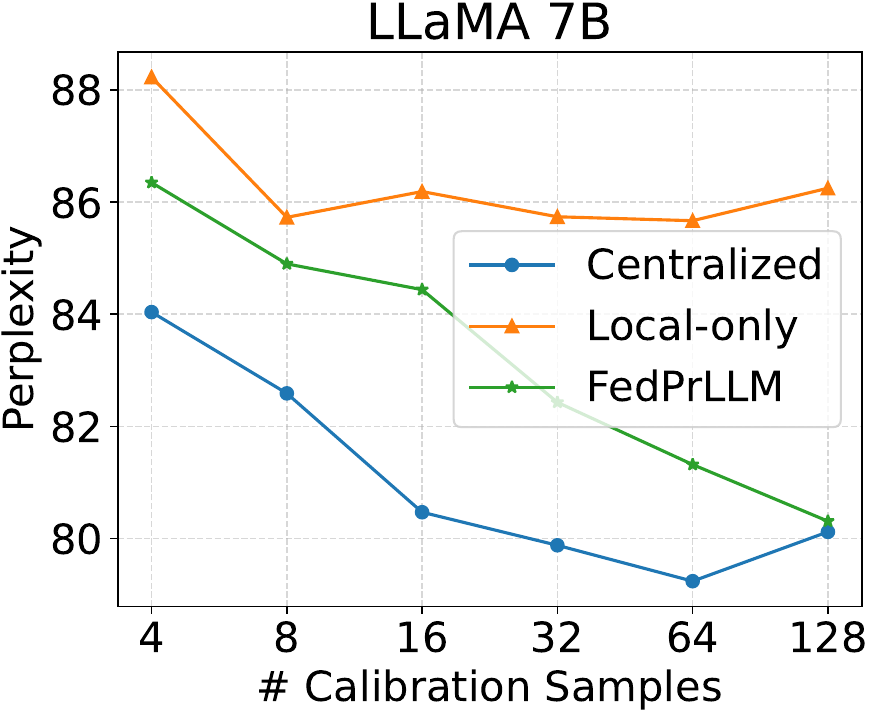}} 
\vspace{-0.1cm}
\caption{The effect of the number of calibration samples on federated pruning LLMs.}
\label{fig:ablation_NS}
\vspace{-0.2cm}
\end{figure}

We further investigate the impact of pruning LLMs in a federated scenario with varying numbers of calibration samples, as shown in Figure \ref{fig:ablation_NS}. Specifically, we change the total number of calibration samples from 128 to 4 while keeping the number of clients equal to half of that. As shown in Figure \ref{fig:ablation_NS}, we observe that with different numbers of calibration samples, FedPrLLM consistently outperforms Local-only pruning, which again shows the effectiveness of the federated pruning method.

\section{Related Work}
\label{sec:related_work}

There is one work that attempts to conduct LLM pruning in an FL scenario \cite{bai2024fedspallm}. Specifically, it enables clients to locally prune their models based on private data and send the pruned models to the server for aggregation. The server averages the pruned models using the FedAvg algorithm \cite{mcmahan2017communication} and prunes the model to satisfy the predefined sparsity rate based on an aggregated mask matrix. This method can be viewed as a specific case within our FedPrLLM framework, i.e., iterative pruning with weight scaling. However, our extensive evaluations reveal that this approach is not optimal.

\section{Conclusion}
\label{sec:conclusion}

In this work, we introduce \textbf{FedPrLLM}, a comprehensive federated pruning framework designed for the privacy-preserving compression of LLMs, incorporating various possibilities for integrating FL with LLM pruning. 
To identify the optimal operation within this framework, we invested thousands of GPU hours exploring these possibilities, including different comparison groups, pruning strategies, and the decision to scale weights. Our extensive evaluation reveals that one-shot pruning with layer comparison and no weight scaling is the optimal choice within the FedPrLLM framework. We hope our work will help guide future efforts in pruning LLMs in privacy-sensitive fields.

\bibliographystyle{plain}
\bibliography{ref}

\clearpage
\appendix

\section{Additional Experimental Results}

\begin{table} [ht]
\centering
\caption{WikiText2 perplexity of pruned LLaMA, LLaMA-2, and LLaMA-3 models.}
\label{app_tab:main_results_wiki}
    \resizebox{\linewidth}{!}{
    \begin{tabular}{cl lcc ccc cc c}
    \toprule
        & & Compar. & Prune & Weight & \multicolumn{3}{c}{LLaMA} & \multicolumn{2}{c}{LLaMA-2} & LLaMA-3 \\
        \cmidrule(lr){6-8} \cmidrule(lr){9-10} \cmidrule(lr){11-11} 
        Sparsity & Method & Group & Stra. & Scaling & 7B & 13B & 30B & 7B & 13B & 8B \\
        \midrule
        0\% & Dense & - & - & - & 5.67 & 5.09 & 4.10 & 5.11 & 4.57 & 7.46 \\
        \midrule
        \multirow{14}{*}{50\%} & Centralized & - & - & - & 7.25 & 6.15 & 5.24 & 6.46 & 5.58 & 11.00 \\
        & Local-only & - & - & - & 7.44 & 6.33 & 5.34 & 6.63 & 5.72 & 11.39 \\
        \cmidrule(lr){2-11}
        & \multirow{12}{*}{FedPrLLM} & Layer & One-shot & \ding{55} & 7.32 & 6.19 & 5.24 & 6.48 & 5.61 & 11.02 \\ 
        & & Row & One-shot & \ding{55} & 7.30 & 6.20 & 5.25 & 6.48 & 5.61 & 11.02 \\
        & & Column & One-shot & \ding{55} & 1524.28 & 9282.09 & 501.88 & 20528.41 & 5309.48 & 311468.53 \\
        & & Layer & Iterative & \ding{55} & 7.30 & 6.19 & 5.24 & 6.48 & 5.62 & 11.12 \\ 
        & & Row & Iterative & \ding{55} & 7.30 & 6.20 & 5.24 & 6.48 & 5.61 & 11.11 \\
        & & Column & Iterative & \ding{55} & 1822.89 & 6884.15 & 996.57 & 77245.84 & 5430.81 & 189134.78 \\
        & & Layer & One-shot & \checkmark & 7.48 & 6.36 & 5.35 & 6.67 & 5.75 & 11.75\\ 
        & & Row & One-shot & \checkmark & 7.47 & 6.36 & 5.35 & 6.67 & 5.75 & 11.75\\
        & & Column & One-shot & \checkmark & 1708.41 & 10819.42 & 824.5 &18084.02 & 5914.91 & 276031.34\\
        & & Layer & Iterative & \checkmark &  7.46 & 6.35 & 5.34 & 6.67 & 5.75 & 11.86\\ 
        & & Row & Iterative & \checkmark &7.46 & 6.35 & 5.34 & 6.67 & 5.74 & 11.87 \\
        & & Column & Iterative & \checkmark &1985.40& 6692.91 & 939.62 & 66911.49 & 5268.71 & 41996.95\\
        \midrule
        \multirow{14}{*}{60\%} & Centralized & - & - & - & 10.71 & 8.74 & 6.55 & 10.03 & 7.92 & 25.81 \\
        & Local-only & - & - & - & 11.70 & 9.38 & 6.96 & 10.84 & 8.55 & 27.47 \\
        \cmidrule(lr){2-11}
        & \multirow{12}{*}{FedPrLLM} & Layer & One-shot & \ding{55} & 10.76 & 8.80 & 6.65 & 10.08 & 8.01 & 25.48 \\ 
        & & Row & One-shot & \ding{55} & 10.77 & 8.80 & 6.64 & 10.08 & 8.03 & 25.64 \\
        & & Column & One-shot & \ding{55} & 2861.56 & 11190.34 & 1047.94 & 14737.65 & 5385.33 & 382319.37 \\
        & & Layer & Iterative & \ding{55} & 10.87 & 8.88 & 6.65 & 10.17 & 8.05 & 26.21\\ 
        & & Row & Iterative & \ding{55} & 10.85 & 8.90 & 6.64 & 10.18 & 8.05 & 25.98\\
        & & Column & Iterative & \ding{55} & 3154.68 & 7824.46 & 2250.97 & 18849.20 & 6556.50 & 65475.84\\
        & & Layer & One-shot & \checkmark & 12.14 & 9.77 & 7.10 & 11.53 & 8.98 & 30.34 \\ 
        & & Row & One-shot & \checkmark & 12.16 & 9.77 & 7.09 & 11.53 & 9.00 & 30.44 \\
        & & Column & One-shot & \checkmark & 3785.85 & 17163.16 & 1770.89 & 15180.33 & 5401.19 & 608169.33 \\
        & & Layer & Iterative & \checkmark & 12.27 & 9.85 & 7.12 & 11.90 & 9.07 & 30.94 \\ 
        & & Row & Iterative & \checkmark & 12.24 & 9.86 & 7.13 & 11.87 & 9.06 & 31.08 \\
        & & Column & Iterative & \checkmark & 2189.53 & 6032.71 & 2626.57 & 16081.73 & 6227.41 & 165510.73 \\
        \midrule
        \multirow{14}{*}{70\%} & Centralized & - & - & - & 87.42 & 53.48 & 17.30 & 72.38 & 45.94 & 92.20 \\
        & Local-only & - & - & - & 104.15 & 67.13 & 23.29 & 80.39 & 51.79 & 108.35 \\
        \cmidrule(lr){2-11}
        & \multirow{12}{*}{FedPrLLM} & Layer & One-shot & \ding{55} & 83.12 & 55.92 & 18.73 & 70.92 & 44.98 & 102.88 \\ 
        & & Row & One-shot & \ding{55} & 81.97 & 56.99 & 18.67 & 70.61 & 44.66 & 102.13 \\
        & & Column & One-shot & \ding{55} & 17281.43 & 13045.16 & 2670.43 & 31238.51 & 12206.74 & 458666.00 \\
        & & Layer & Iterative & \ding{55} & 89.25 & 55.48 & 18.65 & 79.27 & 45.89 & 100.37 \\ 
        & & Row & Iterative & \ding{55} & 92.29 & 57.18 & 18.23 & 72.60 & 45.68 & 93.13  \\
        & & Column & Iterative & \ding{55} & 19791.05 & 10323.63 & 3935.54 & 23090.20 & 7857.41 & 355916.56 \\
        & & Layer & One-shot & \checkmark & 136.50 & 94.90 & 31.62 & 93.89 & 64.34 & 123.92 \\ 
        & & Row & One-shot & \checkmark & 136.09 & 95.86 & 31.48 & 93.36 & 63.98 & 124.65 \\
        & & Column & One-shot & \checkmark & 20505.56 & 11695.06 & 3032.65 & 31485.38 & 10875.86 & 831352.18 \\
        & & Layer & Iterative & \checkmark & 174.95 & 102.78 & 31.12 & 94.49 & 62.07 & 116.97 \\ 
        & & Row & Iterative & \checkmark & 182.73 & 99.32 & 30.87 & 96.37 & 62.51 & 120.19 \\
        & & Column & Iterative & \checkmark & 8607.36 & 11707.00 & 3145.32 & 36254172.00 & 9604.48 & 1034635.56 \\
    \bottomrule
    \end{tabular}
    }
\end{table}

\begin{table} [ht]
\centering
\caption{C4 perplexity of pruned LLaMA, LLaMA-2, and LLaMA-3 models.}
\label{app_tab:main_results_c4}
    \resizebox{\linewidth}{!}{
    \begin{tabular}{cl lcc ccc cc c}
    \toprule
        & & Compar. & Prune & Weight & \multicolumn{3}{c}{LLaMA} & \multicolumn{2}{c}{LLaMA-2} & LLaMA-3 \\
        \cmidrule(lr){6-8} \cmidrule(lr){9-10} \cmidrule(lr){11-11} 
        Sparsity & Method & Group & Stra. & Scaling & 7B & 13B & 30B & 7B & 13B & 8B \\
        \midrule
        0\% & Dense & - & - & - & 7.34 & 6.79 & 6.12 & 7.03 & 6.51 & 12.34 \\
        \midrule
        \multirow{14}{*}{50\%} & Centralized & - & - & - & 9.34 & 8.14 & 7.28 & 8.94 & 8.03 & 18.38  \\
        & Local-only & - & - & - & 9.59 & 8.37 & 7.52 & 9.16 & 8.31 & 18.92 \\
        \cmidrule(lr){2-11}
        & \multirow{12}{*}{FedPrLLM} & Layer & One-shot & \ding{55} & 9.43 & 8.22 & 7.39 & 9.01 & 8.18 & 18.32 \\ 
        & & Row & One-shot & \ding{55} & 9.43 & 8.22 & 7.39 & 9.01 & 8.19 & 18.32 \\
        & & Column & One-shot & \ding{55} & 893.05 & 10616.94 & 612.27 & 9631.37 & 5075.92 & 200257.70\\
        & & Layer & Iterative & \ding{55} & 9.44 & 8.22 & 7.39 & 9.01 & 8.19 & 18.43 \\ 
        & & Row & Iterative & \ding{55} & 9.44 & 8.22 & 7.39 & 9.02 & 8.18 & 18.38 \\
        & & Column & Iterative & \ding{55} & 1050.26	& 8567.66&	779.01	&11658.80&	4804.46	&112192.42\\
        & & Layer & One-shot & \checkmark & 9.64	&8.40	&7.57&	9.21 &	8.39 &	19.45\\ 
        & & Row & One-shot & \checkmark & 9.64&	8.40	&7.57	&9.21&	8.39&	19.45 \\
        & & Column & One-shot & \checkmark &	 887.34 &		13744.66 &895.18	&11440.51&	5189.73 &	90476.94 \\
        & & Layer & Iterative & \checkmark & 9.64 &	8.41 &	7.57 &	9.22 &	8.39 &	19.58\\ 
        & & Row & Iterative & \checkmark & 9.65	&8.41	&7.57&	9.22&	8.39&	19.60\\
        & & Column & Iterative & \checkmark & 1242.31&	6860.69&	724.28	&10355.87&	4657.88	& 44469.52\\
        \midrule
        \multirow{14}{*}{60\%} & Centralized & - & - & - & 13.72 & 11.22 & 9.16 & 13.64 & 11.39 & 43.02 \\
        & Local-only & - & - & - & 14.69 & 11.91 & 9.58 & 14.68 & 12.17 & 45.25 \\
        \cmidrule(lr){2-11}
        & \multirow{12}{*}{FedPrLLM} & Layer & One-shot & \ding{55} & 13.80& 	11.23	&9.29&	13.77&	11.40&	42.61 \\ 
        & & Row & One-shot & \ding{55} & 15.26	&12.24	&9.79&	15.60	&12.75&	50.37 \\
        & & Column & One-shot & \ding{55} & 2149.09	 &11488.68&	993.56	&12252.16	&4606.43	&837570.62 \\
        & & Layer & Iterative & \ding{55} & 13.92	&11.37	&9.32	&13.84&	11.52	&44.24\\ 
        & & Row & Iterative & \ding{55} &13.86&	11.38&	9.30	&13.85	&11.53	&43.77\\
        & & Column & Iterative & \ding{55}& 2981.52&	10375.02&	1752.73	&16673.62&	5289.35	&62234.32 \\
        & & Layer & One-shot & \checkmark &15.24&	12.24&	9.80&	15.61	&12.74&	50.28 \\ 
        & & Row & One-shot & \checkmark &15.26	&12.24	&9.79&	15.60&	12.75&	50.37 \\
        & & Column & One-shot & \checkmark &3336.72	&19430.46&	1520.32&	14613.11&	4547.54&	622715.25 \\
        & & Layer & Iterative & \checkmark &15.46	&12.54&	9.86	&16.15&	13.01&	51.47 \\ 
        & & Row & Iterative & \checkmark & 15.42&	12.54	&9.87	&16.10&	13.01	&51.48\\
        & & Column & Iterative & \checkmark & 1825.82& 	6669.63& 	1865.50& 	16167.12& 	5057.57& 	145341.28 \\
        \midrule
        \multirow{14}{*}{70\%} & Centralized & - & - & - & 85.84 & 53.35 & 18.80 & 84.16 & 58.56 & 136.66  \\
        & Local-only & - & - & - & 96.47 & 63.61 & 22.48 & 82.96 & 67.09 & 161.86 \\
        \cmidrule(lr){2-11}
        & \multirow{12}{*}{FedPrLLM} & Layer & One-shot & \ding{55}&81.95&	52.55&	19.24&	81.40&	59.87&	158.08  \\ 
        & & Row & One-shot & \ding{55}& 82.02&	53.51&	19.22	&81.59&	59.97	&157.87\\
        & & Column & One-shot & \ding{55}& 15276.62	&14041.01&	2059.83	&39339.21	&11306.11	&398674.93 \\
        & & Layer & Iterative & \ding{55} &83.52	&57.22	&19.15&	92.51&	60.46	&162.29 \\ 
        & & Row & Iterative & \ding{55} & 86.77 &	55.98 &	19.20 &	84.99 &	60.86 &	144.71\\
        & & Column & Iterative & \ding{55} & 18149.76 &	13537.18 &	2874.83 &	21704.32 &	7166.78 &	346598.5\\
        & & Layer & One-shot & \checkmark & 116.61	&77.99	&26.30&	104.86	&79.82	&184.11 \\ 
        & & Row & One-shot & \checkmark &117.29&	78.84&	26.29	&104.51&	79.76&	184.11 \\
        & & Column & One-shot & \checkmark &19380.0&	10934.98&	2336.68&	32034.07&	11360.57&	345798.53 \\
        & & Layer & Iterative & \checkmark &142.08&	85.91	&27.17&	103.02	&79.25&	177.97 \\ 
        & & Row & Iterative & \checkmark & 145.15&	84.68 &	27.00 &	102.61 &	79.41 &	182.36 \\
        & & Column & Iterative & \checkmark & 7664.62 &	15985.50&	2685.76&	27805842.0&	8041.09&	1031318.56 \\
    \bottomrule
    \end{tabular}
    }
\end{table}

\begin{table} [ht]
\centering
\caption{PTB perplexity of pruned LLaMA, LLaMA-2, and LLaMA-3 models.}
\label{app_tab:main_results_ptb}
    \resizebox{\linewidth}{!}{
    \begin{tabular}{cl lcc ccc cc c}
    \toprule
        & & Compar. & Prune & Weight & \multicolumn{3}{c}{LLaMA} & \multicolumn{2}{c}{LLaMA-2} & LLaMA-3 \\
        \cmidrule(lr){6-8} \cmidrule(lr){9-10} \cmidrule(lr){11-11} 
        Sparsity & Method & Group & Stra. & Scaling & 7B & 13B & 30B & 7B & 13B & 8B \\
        \midrule
        0\% & Dense & - & - & - & 41.15 & 28.09 & 23.51 & 50.20 & 56.51 & 13.30 \\
        \midrule
        \multirow{14}{*}{50\%} & Centralized & - & - & - & 80.12 & 36.41 & 26.64 & 96.99 & 86.83 & 20.69 \\
        & Local-only & - & - & - & 86.25 & 37.57 & 27.13 & 108.66 & 91.92 & 21.43 \\
        \cmidrule(lr){2-11}
        & \multirow{12}{*}{FedPrLLM} & Layer & One-shot & \ding{55} & 80.31	&36.57	&26.69	&102.71	&88.26	&20.56 \\ 
        & & Row & One-shot & \ding{55}& 80.71	&36.61	&26.64&	101.85	&88.31&	20.55 \\
        & & Column & One-shot & \ding{55} &4463.92	&22138.56	&713.56	&14256.86	&7392.64	&407313.84 \\
        & & Layer & Iterative & \ding{55} &81.22	&36.54	&26.68	&102.72	&88.38	&20.55 \\ 
        & & Row & Iterative & \ding{55} &81.26	&36.55	&26.64	&103.66	&88.94	&20.60\\
        & & Column & Iterative & \ding{55} &4061.96	&17610.52	&1158.75	&13401.63 	&6941.72	&168643.04\\
        & & Layer & One-shot & \checkmark & 87.97&	37.70&	27.27	&112.52&	92.90&	22.21\\ 
        & & Row & One-shot & \checkmark & 88.35	&37.72	&27.25	&112.17	&93.07	&22.21 \\
        & & Column & One-shot & \checkmark &4557.48	&29140.28	&982.59	&12021.08	&7801.23	&264723.12 \\
        & & Layer & Iterative & \checkmark &87.28	&37.69	&27.27&	112.95	&92.58	&22.39\\ 
        & & Row & Iterative & \checkmark &87.61	&37.60	&27.27	&113.32	&92.61	&22.41\\
        & & Column & Iterative & \checkmark & 6929.83	&15189.83	&1178.40	&10208.03	&5220.64	&39172.53\\
        \midrule
        \multirow{14}{*}{60\%} & Centralized & - & - & - & 193.10 & 71.66 & 34.94 & 363.71 & 220.81 & 52.42 \\
        & Local-only & - & - & - & 208.48 & 82.24 & 37.27 & 409.47 & 271.49 & 55.39 \\
        \cmidrule(lr){2-11}
        & \multirow{12}{*}{FedPrLLM} & Layer & One-shot & \ding{55} &187.00	&74.66	&35.38	&339.79	&241.14	&52.61\\ 
        & & Row & One-shot & \ding{55} &186.10	&74.64	&35.47	&337.69	&242.96	&52.61\\
        & & Column & One-shot & \ding{55} &5604.92	&31222.37	&1338.25	&28046.95	&7553.32	&322022.84\\
        & & Layer & Iterative & \ding{55} &191.22	&72.90	&35.83	&368.87	&237.45	&53.78 \\ 
        & & Row & Iterative & \ding{55} &190.60	&73.74	&35.77	&367.56	&235.51	&53.25\\
        & & Column & Iterative & \ding{55} &6785.79	&13234.02	&1903.66	&24022.75	&8125.57	&46139.19\\
        & & Layer & One-shot & \checkmark &216.09	&91.63	&38.22	&429.58	&293.11	&60.49\\ 
        & & Row & One-shot & \checkmark &215.50	&91.60	&38.25	&428.87	&294.44	&60.48\\
        & & Column & One-shot & \checkmark &7600.58	&41079.65	&1910.36	&18249.40	&7601.34	&416094.71\\
        & & Layer & Iterative & \checkmark &220.22	&90.60	&38.79	&427.12	&283.34	&61.25\\ 
        & & Row & Iterative & \checkmark &220.16	&90.58	&38.74	 &428.36	&282.20	&61.55\\
        & & Column & Iterative & \checkmark & 4242.84&	11345.68&	2133.62	&29512.89	&7113.24&	133467.18\\
        \midrule
        \multirow{14}{*}{70\%} & Centralized & - & - & - & 698.79 & 299.42 & 110.70 & 1902.56 & 735.73 & 131.13 \\
        & Local-only & - & - & - & 782.42 & 412.24 & 144.90 & 1780.26 & 863.50 & 152.97 \\
        \cmidrule(lr){2-11}
        & \multirow{12}{*}{FedPrLLM} & Layer & One-shot & \ding{55} &737.07	&366.28	&120.33	&1521.25	&793.55	&156.63 \\ 
        & & Row & One-shot & \ding{55} &718.37	&369.65	&118.24	&1557.08	&792.08	&154.72\\
        & & Column & One-shot & \ding{55} &18649.81	&18136.88	&3180.23	&49646.82	&12010.97	& 466632.84\\
        & & Layer & Iterative & \ding{55} &721.31	&355.21	&113.31	&1675.79	&775.69	&146.27\\ 
        & & Row & Iterative & \ding{55} &734.43	&349.63	&113.65	&1757.10	&767.13 &	133.92\\
        & & Column & Iterative & \ding{55} &28179.23	&17249.42	&3967.48	&29254.5	&10233.18	&314505.62\\
        & & Layer & One-shot & \checkmark &839.42	&484.11	&188.18	&1633.85	&890.27	&174.11\\ 
        & & Row & One-shot & \checkmark &830.33	&483.58	&187.11	&1641.92	&891.27	&172.74\\
        & & Column & One-shot & \checkmark &26556.95	&21627.29	&3383.87&	54429.17	&14951.70	&239612.84 \\
        & & Layer & Iterative & \checkmark &887.36	&469.70	&173.86	&1789.42	&858.48	&162.24\\ 
        & & Row & Iterative & \checkmark &896.85	&454.31	&172.48	&1740.04	&879.50&	168.51\\
        & & Column & Iterative & \checkmark &8660.95	&18472.69	&3246.05	&11427895.00	&8037.55	&738685.56\\
    \bottomrule
    \end{tabular}
    }
\end{table}

\clearpage

\begin{table} [t]
\centering
\caption{Perplexity of pruned LLMs under 50\% sparsity ratio when changing the comparison group for the local pruning method. FedPrLLM adopts one-shot pruning and no weight scaling.}
\label{app_tab:evaluate_compar_group}
    \resizebox{0.99\linewidth}{!}{
    \begin{tabular}{l ccl ccc cc c}
    \toprule
    Local Compar. & & & Compar. & \multicolumn{3}{c}{LLaMA} & \multicolumn{2}{c}{LLaMA-2} & LLaMA-3 \\
        \cmidrule(lr){5-7} \cmidrule(lr){8-9} \cmidrule(lr){10-10} 
    Group & Dataset & Method & Group & 7B & 13B & 30B & 7B & 13B & 8B \\
        \midrule
        \multirow{15}{*}{Layer} & \multirow{5}{*}{WikiText2} & Centralized & - & 7.94&	6.57	&5.47&	7.38	&5.92&	12.04\\ 
        & & Local-only & - &8.16&	6.74	&5.58&	7.56	&6.06&	12.43\\
        \cmidrule(lr){3-10}
        & & \multirow{3}{*}{FedPrLLM} & Layer &7.98	&6.60	&5.48	&7.38	&5.95	&12.09 \\
        & &  & Row &31.85	&10.08	&11.33	&39.07	&124.08	&17.51\\
        & &  & Column &1749.59	&10183.32	&541.62	&25258.16	&5503.91	&336255.96\\
        \cmidrule(lr){2-10}
        & \multirow{5}{*}{C4} & Centralized & - & 10.28	&8.63	&7.59	&10.24	&8.49	&19.18\\ 
        & & Local-only & - &10.56	&8.90	&7.86	&10.52&	8.76	&19.64\\
        \cmidrule(lr){3-10}
        & & \multirow{3}{*}{FedPrLLM} & Layer &10.34	&8.71	&7.72	&10.32	&8.63	&19.09\\
        & &  & Row & 34.90&	12.35&	12.75	&29.79	&207.57	&28.05\\
        & &  & Column  &975.75&	12605.58&	553.85&	13950.23	&4899.58	&129415.62\\
        \cmidrule(lr){2-10}
        & \multirow{5}{*}{PTB} & Centralized & - &92.84&43.47&27.25&306.71&119.17&23.14 \\ 
        & & Local-only & - &99.13&45.34&27.87&338.70&136.88&23.69\\
        \cmidrule(lr){3-10}
        & & \multirow{3}{*}{FedPrLLM} & Layer &91.99&43.59&27.25&305.79&124.27&22.85\\
        & &  & Row &284.19&109.14&110.46&1886.94&480.24&44.71\\
        & &  & Column &3976.21&28144.48&711.16&14131.82&7134.88&293147.84\\
        \midrule
        \multirow{15}{*}{Column} & \multirow{5}{*}{WikiText2} & Centralized & - &8.86&7.68&5.67&10.41&6.38&83.67\\ 
        & & Local-only & - &8.86&7.68&5.67&10.41&6.38&83.67\\
        \cmidrule(lr){3-10}
        & & \multirow{3}{*}{FedPrLLM} & Layer &8.86&7.68&5.67&10.41&6.38&83.67\\
        & &  & Row &138.54&100.80&49.17&764.32&2580.88&400.95\\
        & &  & Column &8.86&7.68&5.67&10.41&6.38&83.67\\
        \cmidrule(lr){2-10}
        & \multirow{5}{*}{C4} & Centralized & - &14.10&11.20&8.06&17.90&9.57&30.88\\ 
        & & Local-only & -&14.10&11.20&8.06&17.90&9.57&30.88\\
        \cmidrule(lr){3-10}
        & & \multirow{3}{*}{FedPrLLM} & Layer &14.10&11.20&8.06&17.90&9.57&30.88\\
        & &  & Row &155.15&87.03&48.19&222.47&5135.37&327.77\\
        & &  & Column &14.10&11.20&8.06&17.90&9.57&30.88 \\
        \cmidrule(lr){2-10}
        & \multirow{5}{*}{PTB} & Centralized & - &108.37&47.17&29.22&4567.49&115.68&240.14 \\ 
        & & Local-only & -&108.37&47.17&29.22&4567.49&115.68&240.14\\
        \cmidrule(lr){3-10}
        & & \multirow{3}{*}{FedPrLLM} & Layer &108.37&47.17&29.22&4567.49&115.68&240.14\\
        & &  & Row &1060.91&394.57&239.91&21323.02&1075.71&928.73\\
        & &  & Column&108.37&47.17&29.22&4567.49&115.68&240.14\\
    \bottomrule
    \end{tabular}
    }
\end{table}







\end{document}